\documentclass{llncs}

\usepackage{amsmath}
\usepackage{dsfont}
\usepackage{makeidx}  
\usepackage{graphicx} 
\usepackage{amsfonts}
\usepackage{amssymb}
\usepackage{multicol}
\usepackage{multirow}
\usepackage{qsymbols}
\usepackage{tabularx}
\usepackage[boxed]{algorithm2e}
\usepackage{enumitem}
\usepackage{pdfpages}
\usepackage{url}
\usepackage{paralist}
\usepackage{subfig}

\newcommand*{\argmax}[1]{\ensuremath{\underset{#1}{\operatorname{argmax}}\hspace{0.1cm}}}

\newcommand{\N}{\ensuremath{\mathbb{N}}} 
\newcommand{\R}{\ensuremath{\mathbb{R}}} 

\begin{document}

\title{Efficient Estimation of $k$ for the Nearest Neighbors Class of Methods\\
- unpublished work from 2011 - }

%
\author{Aleksander Lodwich, Faisal Shafait \and Thomas Breuel}
%
%
%
\institute{
contact info:
\email{aleksander[at]lodwich.net}
}

\maketitle              

\begin{abstract}
The k Nearest Neighbors (kNN) method has received much attention in the past decades, where
some theoretical bounds on its performance were identified and where practical optimizations
were proposed for making it work fairly well in high dimensional spaces and on large datasets.
From countless experiments of the past it became widely accepted that the value 
of $k$ has a significant impact on the performance of this method. However, the efficient 
optimization of this parameter has not received so much attention in literature. 
Today, the most common approach is to cross-validate or bootstrap this value for all values 
in question. This approach forces distances to be recomputed many times, even if efficient 
methods are used. Hence, estimating the optimal $k$ can become expensive even on modern systems. 
Frequently, this circumstance leads to a sparse manual search of $k$. 
In this paper we want to point out that a systematic and thorough estimation of the parameter $k$
can be performed efficiently. The discussed approach relies on large matrices, but we want to argue, 
that in practice a higher space complexity is often much less of a problem than repetitive distance computations.
\end{abstract}

\section{Introduction}

Since the introduction of the k Nearest Neighbor (kNN) method by Fix and Hodges in 1951 \cite{Fix51}
a lot of different variants of it have appeared in order to make it suitable to different scenarios.
The most notable improvements were done in terms of adaptive distance metrics 
\cite{adaptive1}\cite{adaptive2}\cite{adaptive3}, fast access via
space partitioning (packed R* trees \cite{Kame93}, kd-trees\cite{kdtree}, X-trees\cite{xtree}, SPY-TEC \cite{spytec}), 
knowledge base (prototype) pruning (\cite{protoopt1},\cite{protoopt2}) or 
classification based on sensitive distributed data (\cite{distributed1},\cite{distributed2},\cite{distributed3}). 
An overview over the state-of-the-art in nearest neighbor techniques is given in \cite{SurveyKNN}.

The continuing richness of investigation work into nearest neighbor can be explained with
the omnipresence of CBR (Case Based Reasoning) type of problems \cite{Aha98theomnipresence} or just
from the practical point of view of its massive parallelizability or simply populartiy. After all nearest neighbor is 
conceptually easy to understand - many students get to learn the nearest neighbor as the first classifier.

All of the beforehand mentioned advances evolve around 
the question of how to optimize the kNN's distance measure for retrieving. 
The method's apparent laziness might be the reason why 
a fast preoptimization of $k$ has not been paid a lot 
attention to.

The proper choice of $k$ is an important factor for achieving maximum performance of a kNN.
However, as we will show, conventional $k$ optimization via cross-validation or
bootstrapping is slow and promises potential for being sped up. Therefore, we will devote this 
paper to the concept of fast $k$ optimization.


There is work addressing this issue in an alternative 
way by introducing \emph{incremental} kNN classifiers based
on different types of trees. This class of nearest neighbors attempts to eliminate the 
influence of k by choosing the right $k$ ad hoc. 
The rationale behind this method is that for classification 
tasks the exact majority of a specific class label is not 
necessarily interesting. It is only interesting when nearest 
neighbor is used as a density estimator. In other cases it is 
generally enough to feel safe about which class label rules 
the nearest set. This class of nearest neighbor starts by 
polling a minimal amount of nearest neighbors. Then it analyzes 
the labels and if the retrieved collection of labels is indecisive
it will poll more nearest neighbors until it considers the 
collection decisive enough.

Naturally, the method does not scale to small values of $k$, 
because e.g. a $k=1$ will never be indecisive. This is a problem 
because we know from experiments that small $k$ are 
often optimal. Incremental nearest neighbors have their strength in very
large databases where typical queries do not need to compute all relative distances. 
During a lifetime of a large database some distances might not be computed at all.
The lazy distance computation make incremental nearest neighbor methods ideal candidates for
real time tasks operating on large volatile data. However, in a cross validation setup
which needs to compute all distances incremental kNNs cannot play 
out their strengths. Because of the restrictions and intended use 
we exempted incremental methods from investigation in this paper.

We organize this paper in three parts. In the first part we will 
study the options to make the estimation of $k$ as fast as possible, in the second 
part we will experimentally compare the result with the conventional approach
and in the last part we will draw a conclusion.

\section{Stating the Problem}

The kNN is commonly considered a classifier from the area of
supervised learning theory. In this theory
there exists a training function $T$ that delivers 
a model $m$ based on a set of options $o$, a matrix of example 
values $V$ and a vector of labels $l$ of respective size ($m=T(o,V,l)$).
In turn, the model $m$ is used in a classification function $C$ that
is presented with a matrix of new examples $V'$ and its task is
to deliver a vector of new labels $l'$ ($l'=C(m,V')$). $T$ and $C$ must
be related but there is no restriction on what the model can be. 
It can be a set of complex items, a matrix, a vector or 
just a single value, indeed.

From the perspective of a human the purpose of a model is
to make predictions about the future. In order to be able
to do this the human brain requires a simplification of the world.
Only with the simplification of the world to a reduced number
of variables and rules it can compute future states faster than they occur.
In case of kNN the model $m$ consists of the data and of the 
smoothing parameter $k$. This means, that the function $T$ is 
an identity function between model and parameters. This conflicts
with the common notion of a model because the production of models commonly 
implies reduction. However, the collected data already are a reduction
of the world! From a practical point of view they can be considered 
as representative model states of the world and the data in the model 
is used to predict the class variable of a new vector before it 
is actually recorded. This fits perfectly well with the original 
notion of models. However, the model is only fixed, when $k$ is fixed.

According to the framework, fixing the model (getting the right value for $k$) 
is the job of the function $T$. Many training techniques in 
machine learning use optimization strategies developed for real 
numbers and open parameter spaces. This is not suitable for $k$ as it 
is an integer value and has known left and right limits: 
an ideal candidate for full search.

Most frameworks for pattern recognition offer macro optimization
for the remaing model parameters that express themself in the initial training options $o$.
The structural compatibility of the kNN with the pattern recognition
frameworks' macro optimization functions seduces the users 
very often to macro optimize $k$. The consequence of this is that kNN must compute
distances repetitively as it cannot assume that specific vectors
will simply exchange their role between training and testing
in the future. In case of kNN this is exceptionally regrettable.

What does macro optimization mean for the computational complexity?
Here we assume - but without loose of generality - that
the dataset $V$ is of size $n$ ($n$-rows in matrix $V$) and can be 
exactly divided into $f$ equally sized partitions ready to
be rearranged into $f$ different train and test setups.
Since everything is being recomputed the computational 
complexity for this kind of cross-validation of 
$k$ for a brute kNN is $O\left(\hat k\cdot\frac{f-1}{f}\cdot n^2\right)$. 
$\hat k$ is the size of the tested range of $k$ and 
since we are considering full search we accept that $\hat k$ 
depends on the size $n$ of the dataset and the number of folds $f$. 
This means that the $\hat k = n\cdot\frac{f-1}{f}$. 
The scan of nearest sets yields a partial complexity 
of $O(\frac{f-1}{f}\cdot n^2)$. Hence, for
full search the total complexity is 
$O\left(\left( \frac{f-1}{f} \right)^2\cdot n^3  + \frac{f-1}{f}\cdot n^2 \right)$.


In order to reduce this high complexity it is necessary to optimize kNN within
the train function $T$ as it has all necessary information about
the relationships among the examples and the labels. This means
that $T(o=\{k\},V,l)$ should become $T(o=\{i\},V,l)$ where $i$ is a vector
of partition indices of the kind $(0,0,0,...,1,1,1,...,f,f,f...)^T$.

\begin{figure}[ht]
	\centering
	\subfloat[Distance matrix and its computation requiring segments for the cross validation of $k$]{\includegraphics[width=2in]{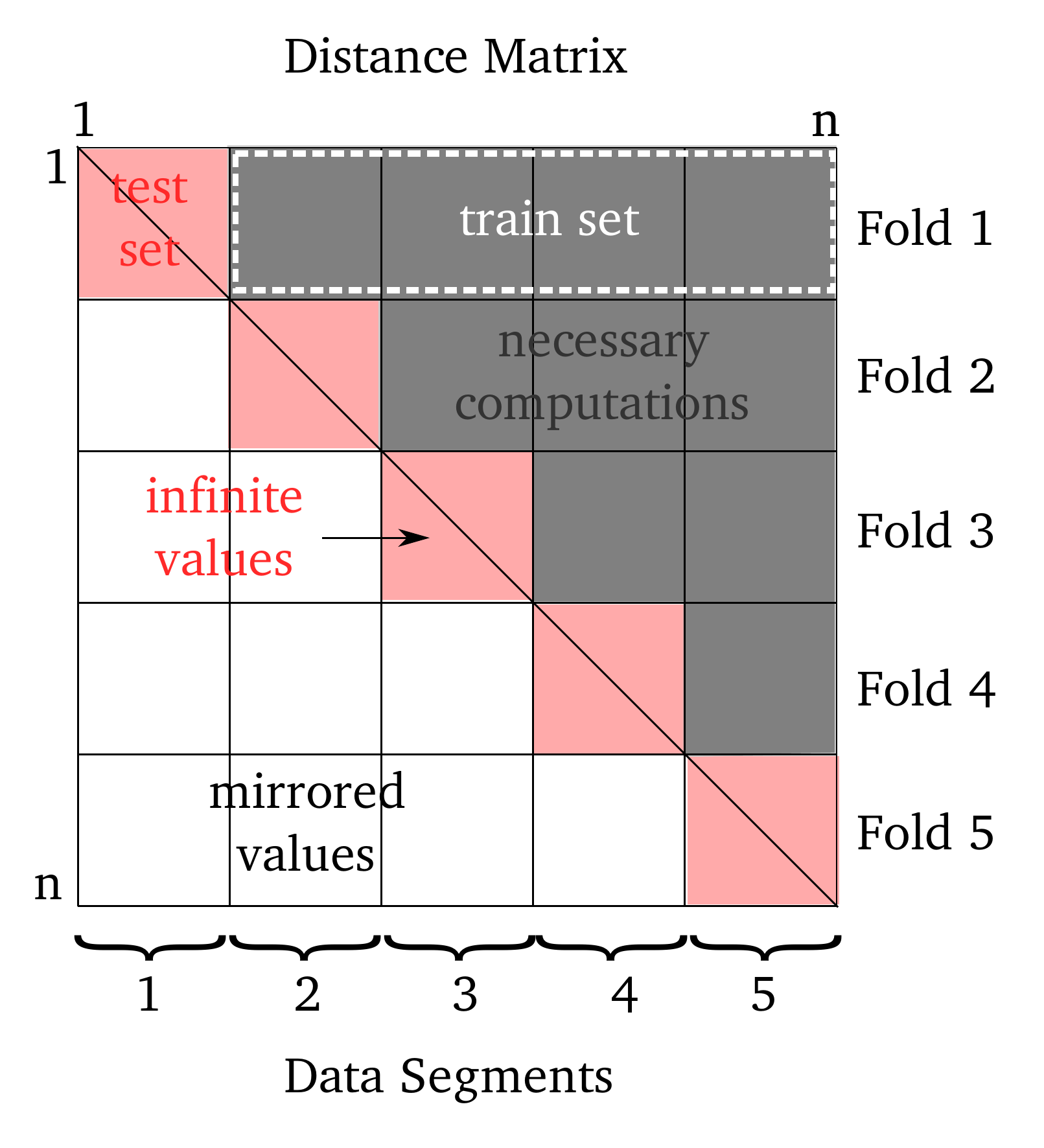}}	\quad
	\subfloat[Sorted distance matrix. Values are being sorted big values to the right]{\includegraphics[width=2.25in]{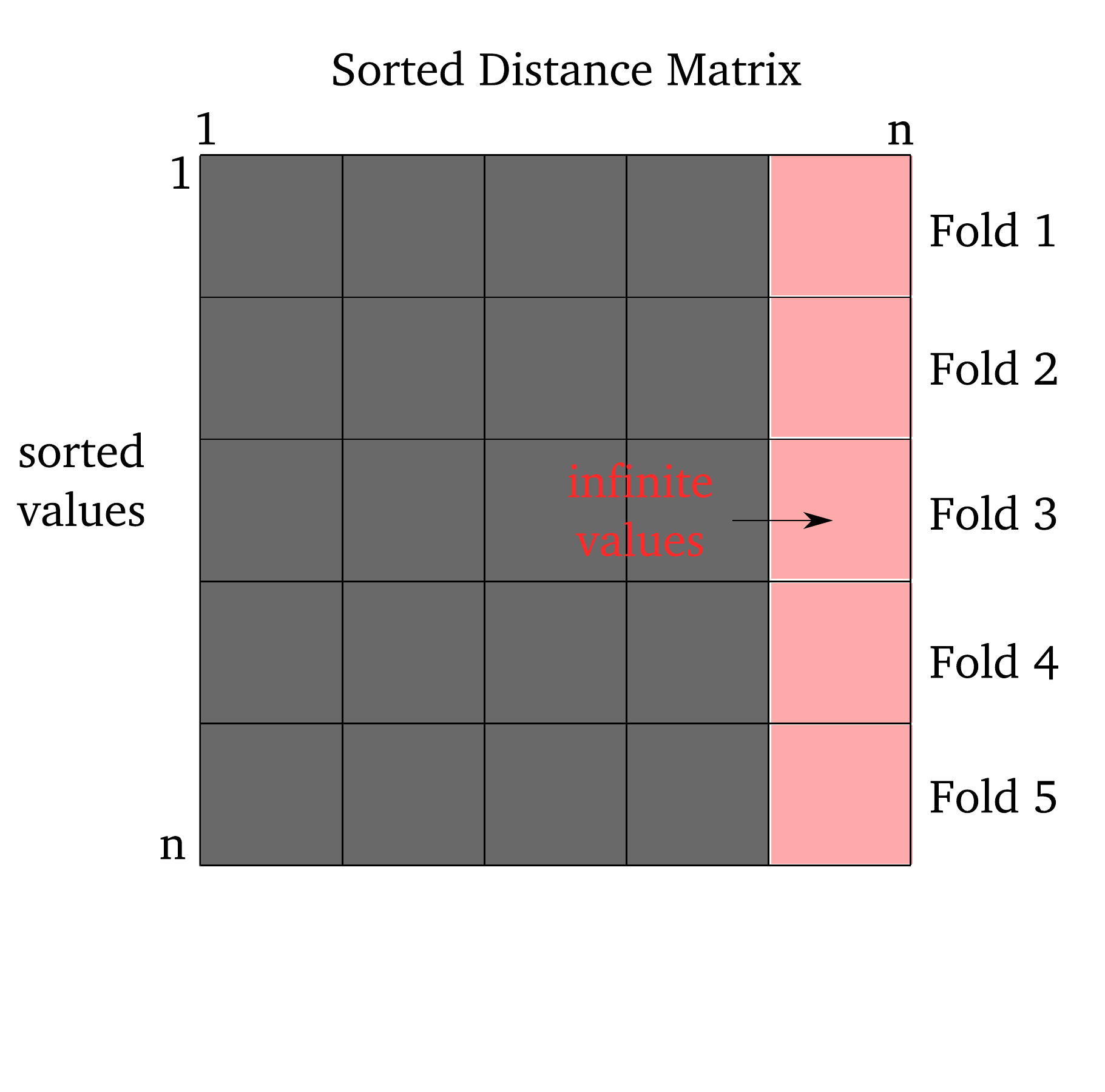}}
	\caption{The distance matrix consists of tuples of data and label ($(d,l)$). 
	It can be sorted horizontally without loosing the relationship 
    between vector pair and ground truth.}

\label{fig:initial_matrix}
\end{figure}

Now, the train function $T$ can utilize the fact that no new
data will arrive during the training and all possible distance requests can
be computed in advance. Distances within the same partition 
need not be computed as they will be never requested. 
These fields can be set to infinity (alternatively they 
can be filled up with the largest value found in the matrix + 1). 
The lower triangle is symmetrical to the upper triangle of 
the matrix because vectors between two points have the same norm. 
The distance matrix $D$ has a structure as shown
in figure \ref{fig:initial_matrix}. The size of $D$ is $n \times n \times 2$.
By $D(column,row)_1$ we mean the component \emph{distance} and 
by $D(column,row)_2$ we mean the associated \emph{label}.

Alone this redesign causes the complexity of the distance computations
to get reduced to $O\left( \frac{f-1}{f}\cdot\frac{n^2}{2} \right)$.
However this benefit is achieved at the expense of higher memory use.
The brute kNN has a space complexity of $O(n)$, now the space 
complexity has risen to $O(n^2)$.

The next step is to sort the vectors horizontally according to 
their distance. Although collecting the $k$ best solutions would 
be faster for a single run it means for a range of $k$ that you 
effectively obtain the insertion sort. Since there exist 
faster sorting algorithms we choose to sort but by using a 
different algorithm. 

The fastest algorithm for doing so is the quick sort. Its average 
complexity is $O\left(n\log(n)\right)$. In the worst case
scenario the sorting complexity of this method is $O\left(n^2\right)$.
In that case $n$ rows will be sorted with $O\left(n^2\right)$.
This means that the worst time complexity so far is
$O\left( \frac{f-1}{f}\cdot\frac{n^2}{2} + n^3\right)$
and average case is $O\left( \frac{f-1}{f}\cdot\frac{n^2}{2} + n^2\log(n)\right)$.

In order to obtain nearest neighbors for each vector indexed
by the row a counting matrix $M:= \N^{n\times s}$ is initialized
with zeros. $s$ is the number of symbols or classes.
For each row in $D$ and $M$ and for the columns 
$k = 1..\frac{f-1}{f}n$ in $D$ the counters for the specific class
label is increased. More precisely, for every row $r = 1..n$
and for every tested $k$ the counters $M\left(D(k,r)_2,r\right)$ 
are updated by $1$. The complexity of this operation
is $O\left(\frac{f-1}{f}\cdot n^2 \right)$. For the overall method
this adds up to $O\left(\frac{3}{2}\frac{f-1}{f}\cdot n^2 \right)$. In parallel the
level of correct classification must be computed because
after every modification of $M$ the state for the smaller
$k$ is lost. Therefore a matrix $A:=\mathbb{N}^{\frac{f-1}{f}n \times f}$ 
for recording the number of correct classfications is required.

How is this number computed? At every round $k$ of the nearest
neighbor candidate computations
$M_k$ contains in each of its rows a vector that tells how many
labels of specific kind are in the nearest neighbors set.
The classification label is $l_{rk}' = \argmax{s} M_{kr}$.
The complexity for this operation is $O( n^2s )$.

This simple method is ambiguous by nature, as there can be many labels
that are represented by the same amount of vectors in a nearby set.
Computer implementations prefer to return the symbol with
the smalest coding. However, it is possible to have a
shadow matrix $S := \R^{n\times s}$ that is the sum of the
distances observed for each class label in the set so far. 
The rule for computing $S$ is the same as for $M$
with the difference that instead of adding ones to the matrix
you add distances. When symbol frequency is ambiguous (argmax
returns more than one value) it is possible to use
$S$ to find which samples are closer overall. Because of
the specific interest into fast $k$ optimization the simple
argmax processing is used.

Now, every $l_{rk}'$ is compared for equality with $l_r$ (ground truth) 
and the binary result is added to $A(k, i_r)$. The $\argmax{k} A_f$
will return $f$ best $k$. $k^*$ is obtained by averaging.

Considering all parts of the algorithm together the overall complexity is
$O\left( \frac{3}{2}\frac{f-1}{f}\cdot n^2 + n^2\log(n) + n^2s \right) \approx O\left( n^2\log(n) \right)$

\begin{figure}[ht]
	\centering
	\subfloat{\includegraphics[width=2.2in]{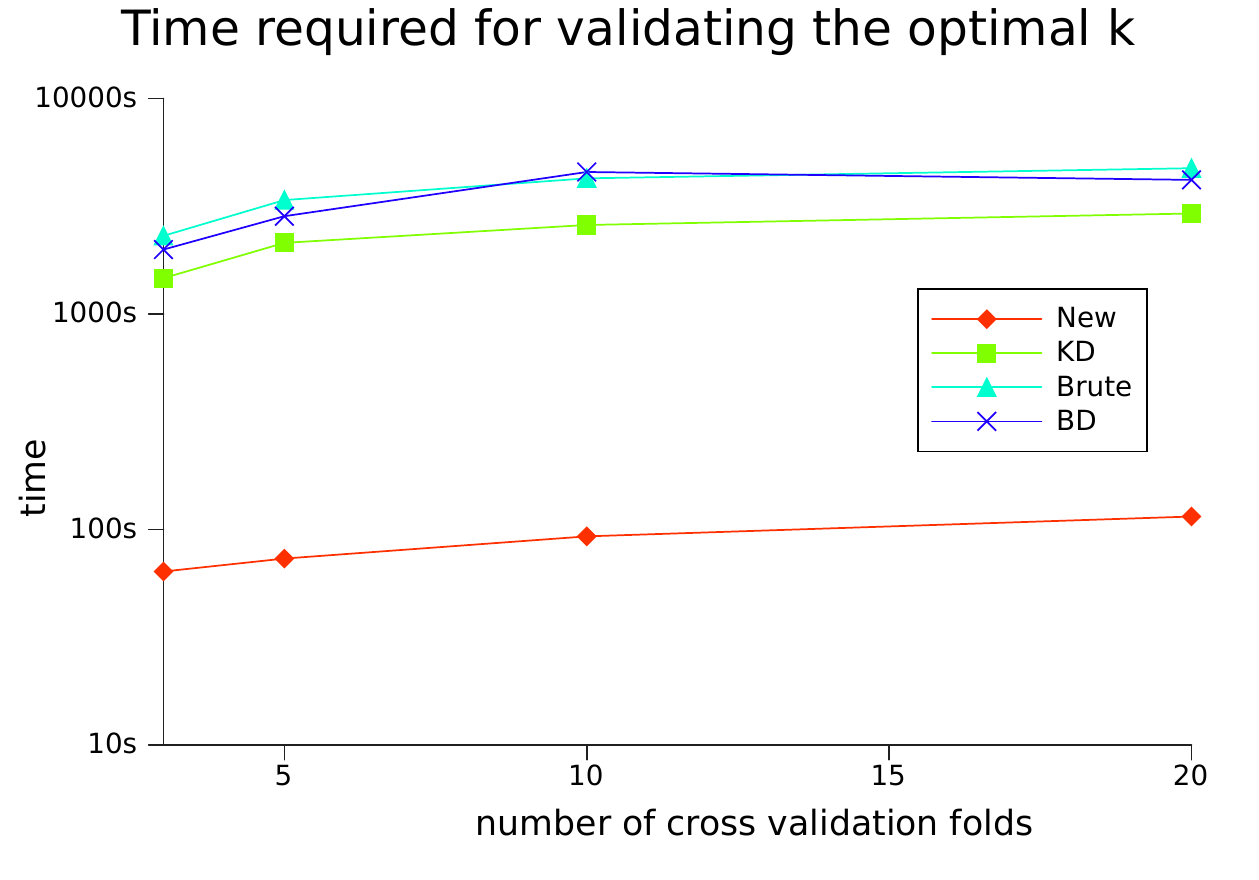}} \quad
	\subfloat{\includegraphics[width=2.2in]{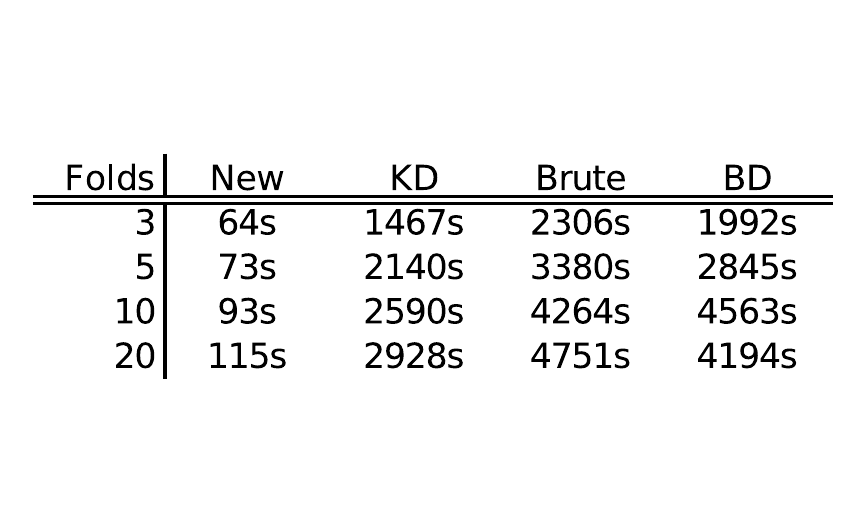}}
	\caption{Experimentally established relationship between the number of folds and the algorithm speed.}
	\label{fig:result}
\end{figure}

\section{Experiments}

The method for fast $k$ computation (AutokNN) was tested against 
three other algorithms from the ANN library 1.1\cite{cit:ann-lib}: brute, kd-tree 
and bd-tree kNN with default settings. The AutokNN and its competitors 
performed a complete cross validation run on the ad, diabetes,
gene, glass, heart, heartc, horse, ionosphere, iris, mushrooms, soybean,
STATLOG australian, STATLOG german, STATLOG heart, STATLOG SAT, STATLOG segment
STATLOG shuttle, STATLOG vehicle, thyroid, waveform and wine datasets
with 3, 5, 10 and 20 folds. The goal of the experiment was the measurement 
of the time required to complete the full course of testing different $k$.

The data was separated into stratified partitions which were used in
different configurations in order to obtain a training and a testing set.
The AutoKNN computes the classification results for all $k$ while the other algorithms
are bound to use a logarithmic search. By logarithmic search the following
schema is meant: $k \in \{1,2,3,4,5,6,7,8,10,100,200,...,1000,...,\frac{f-1}{f}n\}$.
This schema is practically motivated and rational under the assumption that
the influence of additional labels on the result diminishes with higher values
of $k$. Practical consideration is primarily test time. Example: while AutokNN required 15,35s
for a complete scan based on the \emph{ad} dataset, on same data exact brute kNN 
needed 353,7s in logarithmic mode and 19595,7s in full mode. The use of the logarithmic mode
makes results with exactly the same values impossible. However, the 
differences in resulting $k$ and thus in accuracy were absolutely 
negligible so that the results are directly comparable nonetheless.

We added the experiment times for all databases up to a total 
for each cross validation size. The results are shown in the figure 
and the table under \ref{fig:result}. 

Time measurements were performed on a AMD Phenom II 965 with 8GB of RAM with
a Linux 2.6.35 kernel. The algorithms are implemented in C/C++ and 
were compiled with gcc 4.4.5 with O3 option. Only core algorithm operation
was measured and all time for additional I/O was ignored. 
For best comparability, ANN library sources were statically included.

\section{Discussion and Conclusion}
The Nearest Neighbor approach is considered user friendly 
and is frequently used for data mining, classification and regression tasks.
It is embedded into many automatic environments that make
use of kNN's flexibility. Although kNN has been used, analyzed 
and advanced for almost six decades a repeating question can not 
be answered by current literature: What is the fastest way to estimate 
the right value for $k$ and what are the expenses for doing so.

The approach chosen here is to move the $k$ esimation away from the
meta framework right into the training function $T$. The advantage of
this is that additional information about the data can be made. This
additional information allows to precompute the distances among
all vectors without waste and to reuse them numerous times.
From this design change which is known to practitioners but not 
discussed in literature a reduction in time complexity can
be observed from $O\left(\left( \frac{f-1}{f} \right)^2\cdot n^3 + \frac{f-1}{f}\cdot n^2 \right)$
to $O\left( \frac{3}{2}\frac{f-1}{f}\cdot n^2 + n^2\log(n) + n^2s \right)$ in 
average case. The experiments show, that this has significant impact
on the speed of the $k$ estimation task. 
The comparison between kd-tree kNN and the proposed approach proves
moreover that having a better time complexity saves practically
more time than an efficient distance measure for this task.

The cost of this improvement is a higher space complexity (now $O(n^2)$). 
In order to esimtate the practical impact of this complexity exchange 
we studied the contents of the UCI repository \cite{UCI}.
The UCI should be a reasonable crossover of the problems people face in 
real life. 

\begin{figure}[ht]
	\center{\includegraphics[width=4in]{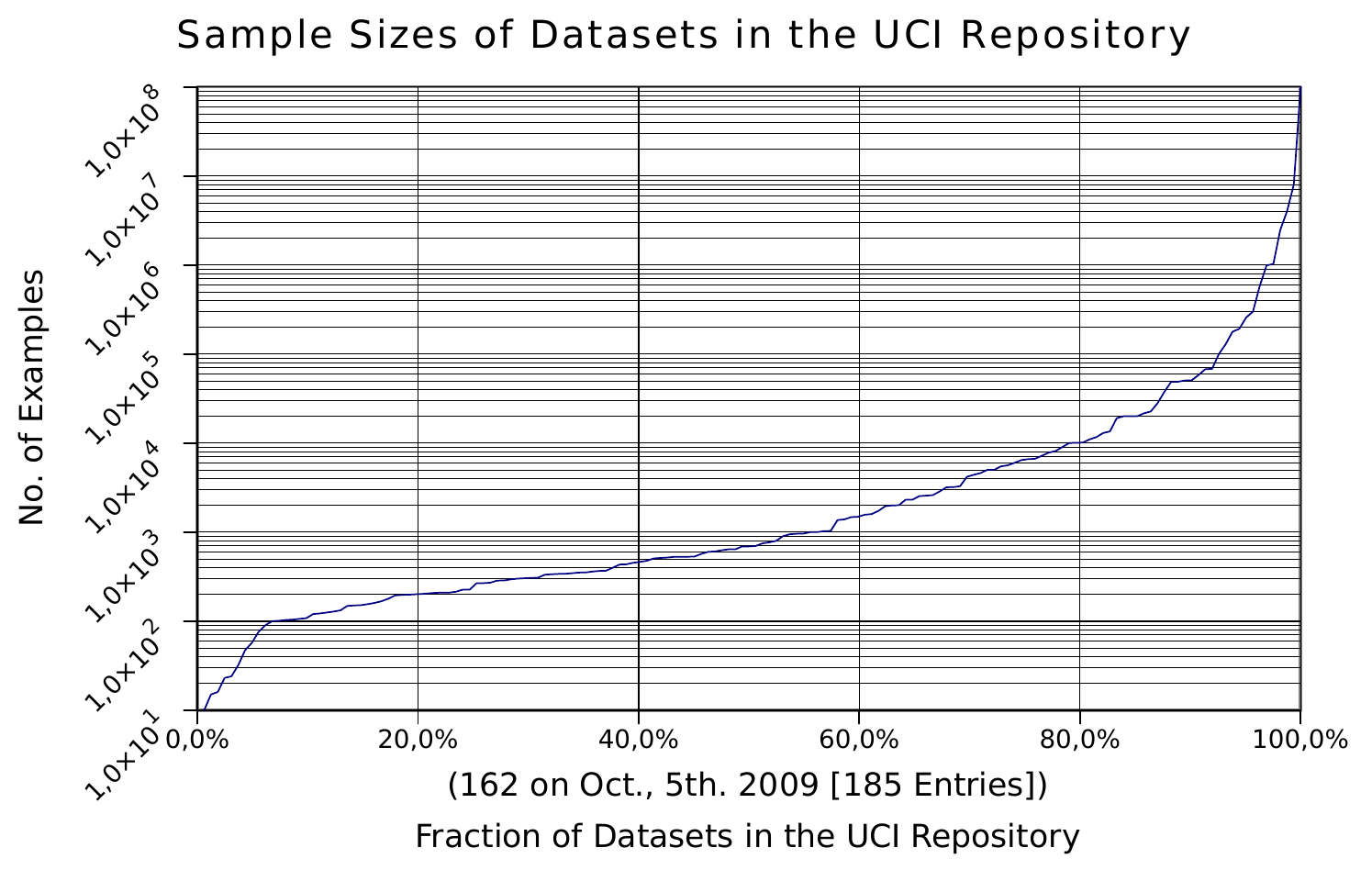}}
	\caption{Dataset sizes in the UCI repository}
	\label{fig:uci}
\end{figure}

Out of 162 datasets we found that 90\% of them have less than 
50K examples, 80\% of them have less than 10K examples and 
half of the UCI's datasets has less than 1000 examples 
(for exact distribution see Fig. \ref{fig:uci}). These sizes
can be easily handled on higher class commodity computers.

This leads to the conclusion that turning in space complexity
for time complexity is a good choice most of the time. Future
implementations should offer an integrated $k$ searching.
The results also show that the so found values for $k$ can be 
transfered not only to other exact kNN but also to 
approximate kNN working on kd-tree and bd-tree models.

\section{Acknowledgments}

This work has been made possible through the funding of the PaREn (Pattern Recognition and Engineering \cite{paren}) project by the BMBF (Federal Ministry of Education and Research, Germany).

%
%

\section*{APPENDIX}
\appendix

\section{Influence of $k$ on the kNN Performance}
\label{apx:perf}
The following diagrams are results from kNN based on natural and synthetic datasets. Synthetic datasets were obtained using WGKS \cite{wgks}
The standard deviation was estimated based on a 10x cross-validation. The diagrams
are non linear. Sections of little change are compressed, hence x-axis are discontinuous.

\includepdf[pages=-]{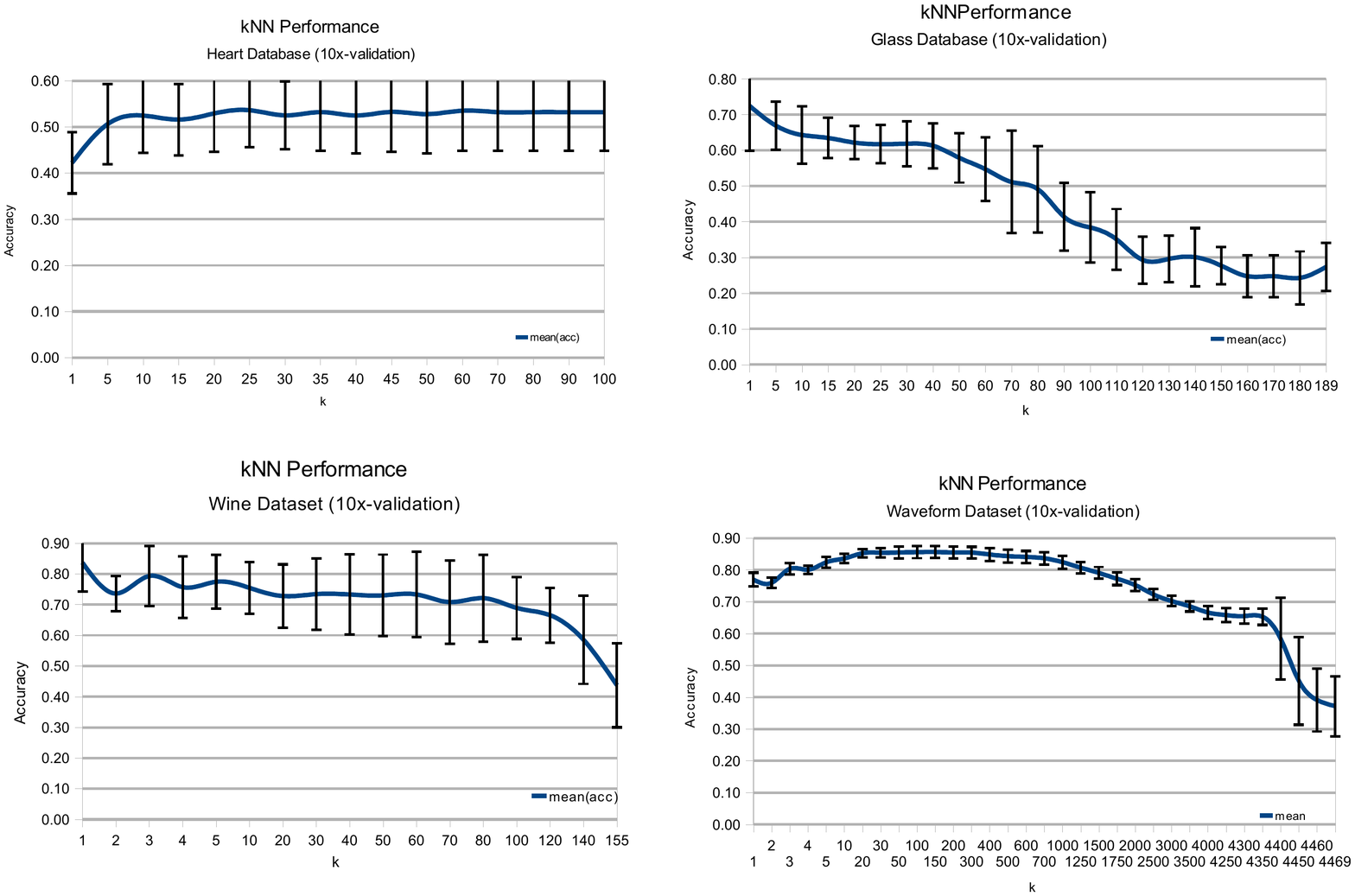}

\end{document}